\documentclass[runningheads]{llncs}

\usepackage{graphicx}
\usepackage{tabu}
\pdfoutput=1
\usepackage{cuted, ragged2e}
\usepackage{marvosym}
\usepackage{tabularray}

\usepackage{tcolorbox}
\tcbuselibrary{most}
\usepackage{array}
\usepackage{hyperref}
\usepackage{stfloats}
\usepackage{amsmath,amssymb,amsfonts}

\usepackage{textcomp}
\usepackage[super]{nth}

\usepackage{xcolor}
\usepackage{booktabs}
\usepackage{multirow}
\usepackage{xspace}
\usepackage{framed}
\usepackage{tablefootnote}
\usepackage[ruled,vlined]{algorithm2e}
\usepackage[noend]{algpseudocode}
\usepackage{subcaption}
\usepackage{bbm}
\usepackage{xspace}
\usepackage{balance}
\usepackage{fancyhdr}

\algnewcommand\algorithmicforeach{\textbf{for each}}
\algdef{S}[FOR]{ForEach}[1]{\algorithmicforeach\ #1\ \algorithmicdo}

\usepackage{tcolorbox}
\usepackage{verbatim}
\usepackage{marginnote}

\usepackage{booktabs}
\newcommand{\lab}[1]{\textquotesingle{#1}\textquotesingle}

\usepackage{xspace}

\newcommand{\qnnrepair}{{\sc QNNRepair}\xspace}
\newcommand{\commentout}[1]{}
\usepackage{marginnote}
\usepackage{color}

\usepackage{soul}

\begin{document}

\title{\qnnrepair: Quantized Neural Network Repair}

\author{Xidan Song\inst{1}\orcidID{0000-0003-2612-6296} \and
Youcheng Sun\inst{1}\orcidID{0000-0002-1893-6259} \and
Mustafa A. Mustafa\inst{1,2}\orcidID{0000-0002-8772-8023} \and
Lucas C. Cordeiro\inst{1,3}\orcidID{0000-0002-6235-4272}}
%
%
\institute{Department of Computer Science, The University of Manchester, UK \\ \and
COSIC, KU Leuven, Belgium \and
Federal University of Amazonas, Brazil \\ \email{\{xidan.song, youcheng.sun, mustafa.mustafa, lucas.cordeiro\}@manchester.ac.uk}}

\maketitle
\thispagestyle{fancy}

\lhead{} 
\chead{} 
\rhead{} 
\lfoot{} 
\cfoot{} 
\rfoot{\thepage} 
\renewcommand{\headrulewidth}{0pt} 
\renewcommand{\footrulewidth}{0pt} 
\pagestyle{fancy}
\rfoot{\thepage}
\begin{abstract}

We present QNNRepair, the first method in the literature for repairing quantized neural networks (QNNs). QNNRepair aims to improve the accuracy of a neural network model after quantization. It accepts the full-precision and weight-quantized neural networks, together with a repair dataset of passing and failing tests. At first, QNNRepair applies a software fault localization method to identify the neurons that cause performance degradation during neural network quantization. Then, it formulates the repair problem into a MILP, solving neuron weight parameters, which corrects the QNN’s performance on failing tests while not compromising its performance on passing tests. We evaluate QNNRepair with widely used neural network architectures such as MobileNetV2, ResNet, and VGGNet on popular datasets, including high-resolution images. We also compare QNNRepair with the state-of-the-art data-free quantization method SQuant~\cite{guo2022squant}. According to the experiment results, we conclude that QNNRepair is effective in improving the quantized model’s performance in most cases. Its repaired models have 24\% higher accuracy than SQuant’s in the independent validation set, especially for the ImageNet dataset.

\keywords{neural network repair, quantization, fault localization, constraints solving}
\end{abstract}

\section{Introduction}
\label{sec:introduction}



Nowadays, neural networks are often used in safety-critical applications, such as autonomous driving, medical diagnosis, and aerospace systems~\cite{zhang2020testing}. In such applications, often quantized (instead of full precision) neural network models are deployed due to the limited computational and memory resources of embedded devices~\cite{guo2018survey}. Since the consequences of a malfunction/error in such applications can be catastrophic, it is crucial to ensure that the network behaves correctly and reliably~\cite{song2020improving}.

Quantized neural networks~\cite{guo2018survey} use low-precision data types, such as 8-bit integers, to represent the weights and activations of the network. While this reduces the memory and computation requirements of the network, it can also lead to a loss of accuracy and the introduction of errors in the network's output. Therefore, it is important to verify that the quantization process has not introduced any significant errors that could affect the safety or reliability of the network.

To limit the inaccuracy in a specific range, various neural network model verification methods~\cite{ehlers2017formal}~\cite{katz2019marabou}~\cite{katz2017reluplex}~\cite{wang2021beta}~\cite{henzinger2021scalable} have been proposed. Neural network verification~\cite{huang2017safety,abs-2106-05997,abs-2111-13110} aims to provide formal guarantees about the behavior of a neural network, ensuring that it meets specific safety and performance requirements under all possible input conditions. They set constraints and properties of the network input and output to check whether the model satisfies the safety properties. However, neural network verification can be computationally expensive, especially for large, deep networks with millions of parameters. This can make it challenging to scale the verification process to more complex models. 
While the majority of the neural network verification work is on full precision models, many verification techniques focus on quantized models as well~\cite{henzinger2021scalable,zhang2022qvip,giacobbe2020many,amir2021smt}.

Other researchers improve the performance and robustness of the trained neural network models by repair~\cite{yu2021deeprepair}~\cite{goldberger2020minimal}~\cite{usman2021nn}~\cite{sotoudeh2021provable}~\cite{borkar2019deepcorrect}. These methods can be divided into three categories: retraining/refining, direct weight modification, and attaching repairing units.
There are also quantized aware training (QAT) techniques~\cite{li2021brecq}~\cite{gong2019differentiable}~\cite{choi2018pact}, a method to train neural networks with lower precision weights and activations, typically INT8 format. QAT emulates the effects of quantization during the training process. 
QAT requires additional steps, such as quantization-aware back-propagation and quantization-aware weight initialization, making the training process more complex and time-consuming. Quantized-aware training methods require datasets for retraining, which consume a lot of time and storage. However, for Data-free quantization like SQuant~\cite{guo2022squant}, which does not require datasets, the accuracy after quantization is relatively low.

In \qnnrepair, we use the well-established software fault localization methods to identify suspicious neurons in a quantized model corresponding to the performance degradation after quantization. 
We then correct these most suspicious neurons' behavior by MILP, in which the constraints are encoded by observing the difference between the quantized model and the original model when inputs are the same. 
The main contributions of this paper are 
three-fold:
\begin{itemize}

\item We propose, implement and evaluate \qnnrepair~-- a new method for repairing QNNs. It converts quantized neural network repair into a MILP (Mixed Integer Linear Programming) problem. \qnnrepair features direct weight modification and does not require the training dataset.

\item We compare \qnnrepair with a state-of-the-art QNN repair method -- Squant~\cite{guo2022squant}, and demonstrate that \qnnrepair can achieve higher accuracy than Squant after repair. We also evaluate \qnnrepair on multiple widely used neural network architectures to demonstrate its effectiveness.

\item We have made \qnnrepair and its benchmark publicly available at: 

\url{https://github.com/HymnOfLight/QNNRepair}

\end{itemize}


\section{Related Work}
\label{sec:RelatedWorks}

\subsection{Neural Network Verification}
\label{sec:Quantized_verification}

The first applicable methods supporting the non-linear activation function for neural network verification can be traced back to 2017, R Ehlers et al.~\cite{ehlers2017formal} proposed the first practicable neural network verification method based on SAT solver (solve the Boolean satisfiability problem)~\cite{een2004extensible}. They present an approach to verify neural networks with piece-wise linear activation functions. Guy Katz et al.~\cite{katz2019marabou} presented Marabou, an SMT(Satisfiability modulo theories)~\cite{de2008z3}-based tool that can answer queries about a network’s properties by transforming these queries into constraint satisfaction problems. However, implementing SMT-based neural network verification tools is limited due to the search space and the scale of a large neural network model, which usually contains millions of parameters~\cite{gehr2018ai2}. The SMT-based neural network verification has also been proved as an NP-complete problem~\cite{katz2017reluplex}. Shiqi Wang et al. develop $\beta$-CROWN~\cite{wang2021beta}, a new bound propagation-based method that can fully encode neuron splits via optimizable parameters $\beta$ constructed from either primal or dual space. Their algorithm is empowered by the $\alpha$,$\beta$-CROWN (alpha-beta-CROWN) verifier, the winning tool in VNN-COMP 2021~\cite{bak2021second}. There are also some quantized neural network verification methods. TA Henzinger et al.~\cite{henzinger2021scalable} proposed a scalable quantized neural network verification method based on abstract interpretation. However, due to the search-space explosion, it has been proved that SMT-based quantized neural network verification is a PSPACE-hard problem~\cite{henzinger2021scalable}.

\subsection{Neural Network Repair}
\label{sec:Repairing}

Many researchers have proposed their full-precision neural network repairing techniques. These can be divided into three categories: Retraining, direct weight modification, and attaching repairing units. 

In the first category of repair methods, the idea is to retrain or fine-tune the model for the corrected output with the identified misclassified input. DeepRepair~\cite{yu2021deeprepair} implements transfer-based data augmentation to enlarge the training dataset before fine-tuning the models. The second category uses solvers to get the corrected weights and modify the weight in the trained model directly. These types of methods, including~\cite{goldberger2020minimal} and~\cite{usman2021nn}, used SMT solvers for solving the weight modification needed at the output
layer for the neural network to meet specific requirements without any retraining. The third category of methods repairs the models by introducing more weight
parameters or repair units to facilitate more efficient repair. PRDNN~\cite{sotoudeh2021provable} introduces a new DNN architecture that enables efficient and effective repair, while DeepCorrect~\cite{borkar2019deepcorrect} corrects the worst distortion-affected filter activations by appending correction units. AIRepair \cite{song2022airepair} aims to integrate multiple existing repair techniques into the same platform.
However, these methods only support the full-precision models and cannot apply to quantized models.

\subsection{Quantized Aware Training}
\label{sec:quantized_aware}

Some researchers use quantized-aware training to improve the performance of the quantized models. Yuhang Li et al. proposed a post-training quantization framework by analyzing the second-order error called BRECQ(Block Reconstruction Quantization)~\cite{li2021brecq}. Ruihao Gong et al. proposed Differentiable Soft Quantization (DSQ)~\cite{gong2019differentiable} to bridge the gap between the full-precision and low-bit networks. It can automatically evolve during training to gradually approximate the standard quantization. J Choi et al.~\cite{choi2018pact}proposed a novel quantization scheme PACT(PArameterized Clipping acTivation) for activations during training - that enables neural networks to work well with ultra-low precision weights and activations without any significant accuracy degradation. However, these methods require retraining and the whole dataset, which will consume lots of computing power and time to improve tiny accuracy in actual practices. In addition, there is a method called Data-free quantization, which quantizes the neural network model without any datasets. Cong Guo et al. proposed SQuant~\cite{guo2022squant}, which can quantize networks on inference-only devices with low computation and memory requirements.

\section{Preliminaries}
\label{sec:Preliminaries}

\subsection{Statistical Fault Localization}

Statistical fault localization techniques (SFL)~\cite{naish2011model} have been widely used in
software testing to aid in locating the causes of failures of programs. During the execution of each test case, data is collected indicating the executed statements. Additionally, each
test case is classified as passed or failed.

This technique uses information about the program's execution traces and associated outcomes (pass/fail) to identify suspicious program statements. It calculates four suspiciousness scores for each statement based on the correlation between its execution and the observed failures.  We use the notation $ C^{\mathrm{af}}, C^{\mathrm{nf}}, C^{\mathrm{as}}, C^{\mathrm{ns}}$. The first part of the superscript indicates whether the statement was executed/``activated'' (a) or not (n), and the
second indicates whether the test is a passing/successful (s) or failing (f) one. For example, $C^{\mathrm{as}}$ is the number of successful tests that execute a statement $C$. Statements with higher suspiciousness scores are more likely to contain faults. There are
many possible metrics that have been proposed in the
literature. We use Tarantula~\cite{jones2005empirical}, Ochiai~\cite{abreu2007accuracy}, DStar~\cite{wong2013dstar}, Jaccard~\cite{agarwal2014fault}, Ample~\cite{dallmeier2005lightweight}, Euclid~\cite{galijasevic2002fault} and Wong3~\cite{wong2007effective}, which are widely used and accepted in the application of Statistical fault localization, in our ranking procedure. We discuss the definition and application in our method of these metrics in Section~\ref{ranking_importance}. In addition, SFL has been also used for analyzing and explaining neural networks~\cite{eniser2019deepfault,sun2020explaining}.

\subsection{Neural Network and Quantization}
\label{sec:repair_definition}

A neural network consists of an input layer, an output layer, and one or more intermediate layers called hidden layers. Each layer is a collection of nodes, called neurons. Each neuron is connected to other neurons by one or more directed edges~\cite{elboher2020abstraction}. 

Let $f: \mathcal{I} \rightarrow \mathcal{O}$ be the neural network $N$ with $m$ layers. In this paper, we focus on a neural network for image classification. For a given input x $\in \mathcal{I}, f(x) \in \mathcal{O}$ calculates the output of the DNN, which is the classification label of the input image.  Specifically, we have

\begin{equation}
\begin{aligned}
f(x)=f_{N}\left(\ldots f_{2}\left(f_{1}\left(x ; W_{1}, b_{1}\right) ; W_{2}, b_{2}\right) \ldots ; W_{N}, b_{N}\right)
\end{aligned}
\end{equation}

In this equation, $W_{i}$ and $b_{i}$ for $i=1,2, \ldots, N$ represent the weights and bias of the model, which are trainable parameters. $f_{i}\left(z_{i-1} ; W_{i-1}, b_{i-1}\right)
$ is the layer function that maps the output of layer $(i-1) \text {, i.e., } z_{i-1}$, to the input layer $i$.

\paragraph{Quantization} As one of the general neural network model optimization methods, model quantization can reduce the size and model inference time of DNN models and their application to most models and different hardware devices. By reducing the number of bits per weight and activation, the model's storage requirements and computational complexity can be significantly optimized. Jacob et al.~\cite{jacob2018quantization} report benchmark results on popular ARM CPUs for state-of-the-art MobileNet architectures, as well as other tasks, showing significant improvements in the latency-vs-accuracy tradeoffs. In the following formula, $r$ is the true floating point value, $q$ is the quantized fixed point value, $Z$ is the quantized fixed point value corresponding to the $0$ floating point value, and $S$ is the smallest scale that can be represented after quantization of the fixed point. The formula for quantization from floating point to fixed point is as follows:

\begin{equation}
\begin{array}{c}
r=S(q-Z) \\
q=\operatorname{round}\left(\frac{r}{S}+Z\right)
\end{array}
\end{equation}

Currently, Google's TensorFlow Lite~\cite{david2021tensorflow} and NVIDIA's TensorRT~\cite{vanholder2016efficient} support the INT8 engine framework. 
\subsection{Solvers for Mixed Integer Linear Optimization}
\label{sec:LP__solver}

MILP (Mixed Integer Linear Programming) is an extension of linear programming in which some or all of the decision variables are restricted to integers. In this type of problem, the objective function and all constraints are linear, but due to the presence of integer constraints, the solution space becomes discrete, making the problem more complex and challenging.

All state-of-the-art solvers for MILP employ one of many existing variants of the well-known branch-and-bound algorithm of~\cite{land2010automatic}. This class of algorithm searches a dynamically constructed tree
(known as the search tree).

The state-of-the-art MILP solvers include Gurobi~\cite{optimization2012inc}, which is a commercial solver widely used for linear programming, integer programming, and mixed integer linear programming. According to B. Meindl and M. Templ's~\cite{meindl2012analysis} Analysis of commercial and free and open source solvers for linear optimization problems, Gurobi is the fastest solver and can solve the largest number of problems. Another reason for choosing Gurobi was primarily in the area of neural network robustness, and other approaches, such as alpha-beta-crown in the area of neural network verification, use Gurobi as their backend. Hence we use Gurobi as the backend to solve the neural network repairing problem.

Other MILP solver include: CPLEX~\cite{nickel2022ibm}, GLPK (GNU Linear Programming Kit)~\cite{makhorin2008glpk}. Python external library Scipy~\cite{bressert2012scipy} also provides some functions for MILP.

\section{\qnnrepair Methodology}
\label{sec:Methodology}

\begin{figure}
    \centering
        \centering
        \includegraphics[width=0.8\textwidth]{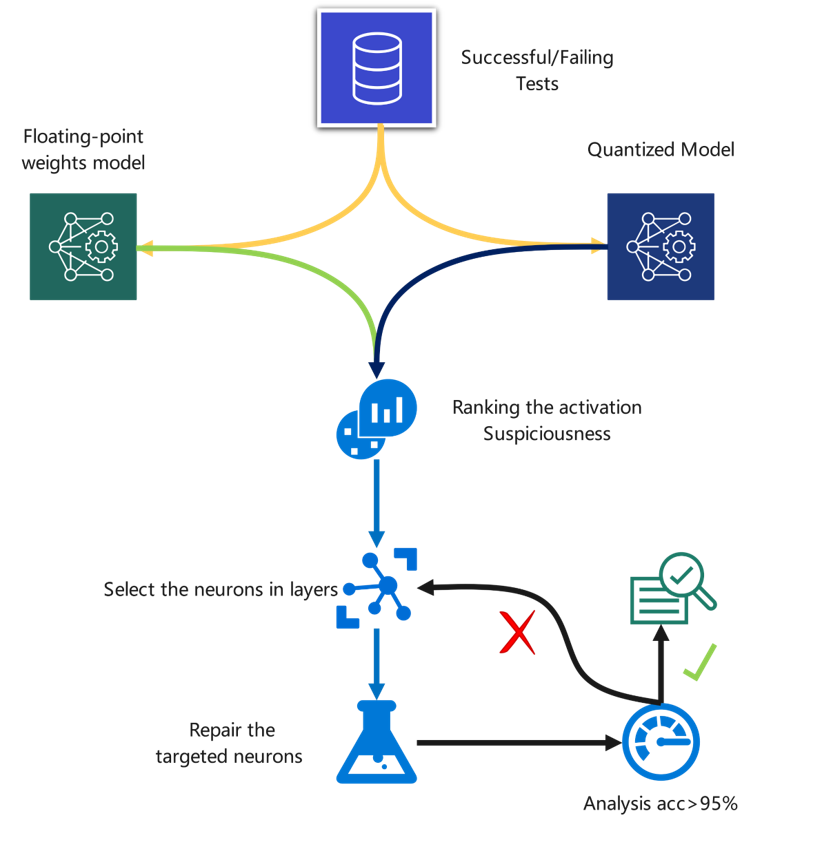}
    \caption{The QNNRepair Architecture.}
    \label{fig:workflow}
\end{figure}

The overall workflow of \qnnrepair is illustrated in Figure~\ref{fig:workflow}. It takes two neural networks, a floating-point model and its quantized version for repair, as inputs.
There is also a repair dataset of successful (passing) and failing tests, signifying whether the two models would produce the same classification outcome when given the same test input.

The passing/failing tests are used by \qnnrepair to evaluate each neuron's importance and localize 
these neurons to repair for improving the quantized model's performance (Section~\ref{ranking_importance}). The test cases can be generated by by dataset augmentation~\cite{shorten2019survey} or various neural network testing methods~\cite{pei2017deepxplore,sun2018concolic,sun2019structural,odena2019tensorfuzz}. 
In \qnnrepair, the neural network repair problem is encoded into a Mixed Integer Linear Programming problem for solving the corrected neuron weights (Section~\ref{sec:repair_constraints}). 
It then replaces the weights with corrected weights, 
which \qnnrepair evaluates the performance of the quantized model by testing its classification accuracy. If the quantized model's performance is good enough w.r.t. the floating point one after repair, the model is ready for deployment. Otherwise,  \qnnrepair continues by selecting other parameters to repair. 
More detailed information is presented in Algorithm~\ref{alg:Repair} (Section~\ref{sec:algorithm}).

\subsection{Ranking the Importance of the Neurons}
\label{ranking_importance}

\qnnrepair starts with evaluating the importance of the neurons in the neural network for causing the output difference between the quantized model and the floating point one. 
When conducting an inference procedure on an image, the intermediate layer in the model has a series of outputs as the inputs for the next layer. The outputs go through activation functions, and we assume it is a ReLU function. For the output, if it is positive, we place it as one. If not, we place it as zero, naming it the activation output. Let $f_{i}$ and $q_{i}$ represent the activation output of a single neuron in full-precision and quantized models separately. If there is a testing image that makes $\left(f_{i}, q_{i}\right)$ not equal, we consider the neuron as ``activated", and we set $v_{m n}=1$, otherwise $v_{m n}=0$. Then we define the activation function matrix to assemble the activation status of all neurons for the floating-point model: 

$\left(\begin{array}{ccc}
f_{11} & \cdots & f_{1 n} \\
\vdots & \ddots & \vdots \\
f_{m 1} & \cdots & f_{m n}
\end{array}\right)=f_{i}$
and $q_{i}$ for the quantized model.

We define the activation differential matrix to evaluate the activation difference between the floating point and the quantized model. Given an input image $i$, we calculate $\operatorname{diff} _{i}=f_{i}-q_{i}$ between the two models. We form a large matrix of these ${diff} _{i}$ regarding the image $i$. The element in this matrix should be $0$ or $1$, representing whether the floating and quantized neural networks’ activation status is the same.

We borrow the concepts from traditional software engineering, just replacing the statements in traditional software with neurons in neural network models. We define the passing tests as the images in the repair set that the floating-point and quantized model have the same classification output, and failing tests as their classification results are different. For a set of repair images, we define $<C_{n}^{\mathrm{af}}, C_{n}^{\mathrm{nf}}, C_{n}^{\mathrm{as}}, C_{n}^{\mathrm{ns}}>$ as following:

\begin{itemize}
    \item {\small $C_{n}^{\mathrm{af}}$} is the number of ``activated" neurons for failing tests.
    \item {\small $C_{n}^{\mathrm{nf}}$} is the number of ``not activated" neurons for failing tests.
    \item {\small $C_{n}^{\mathrm{as}}$} is the number of ``activated" neurons for passing tests.
    \item {\small $C_{n}^{\mathrm{ns}}$} is the number of ``not activated" neurons for passing tests.
\end{itemize}

We borrow the concepts from traditional software fault localization: Tarantula~\cite{jones2005empirical}, Ochiai~\cite{abreu2007accuracy}, DStar~\cite{wong2013dstar}, Jaccard~\cite{agarwal2014fault}, Ample~\cite{dallmeier2005lightweight}, Euclid~\cite{galijasevic2002fault} and Wong3~\cite{wong2007effective} and defined the indicators of neuronal suspicion in Table~\ref{table:importance_metrics_definition}. Note that in DStar, * represents the $n$ square of $C_{n}^{a f}$.

\begin{table*}[t]
\small
\centering
\caption{Importance (i.e., fault localization) metrics used in experiments}
\label{table:importance_metrics_definition}
\resizebox{0.7\linewidth}{!}{
\begin{tabular}{@{\extracolsep{10pt}}llll@{}}
\toprule
Tarantula:          & $\frac{C_{n}^{\mathrm{af}} /\left(C_{n}^{\mathrm{af}}+C_{n}^{\mathrm{nf}}\right)}{\mathrm{C}_{n}^{\mathrm{af}} /\left(C_{n}^{\mathrm{af}}+C_{n}^{\mathrm{nf}}\right)+C_{n}^{\mathrm{as}} /\left(C_{n}^{\mathrm{as}}+C_{n}^{\mathrm{ns}}\right)}$ & Euclid:             & $\sqrt{C_{n}^{\mathrm{af}}+C_{n}^{\mathrm{ns}}}$                          \\ \\
Ochiai:             & $\frac{C_{n}^{\mathrm{af}}}{\sqrt{\left(C_{n}^{\mathrm{af}}+C_{n}^{\mathrm{as}}\right)\left(C_{n}^{\mathrm{af}}+C_{n}^{\mathrm{nf}}\right)}}$                                                                                                    & DStar:              & $\frac{C_{n}^{\mathrm{af}^{*}}}{C_{n}^{\mathrm{as}}+C_{n}^{\mathrm{nf}}}$ \\ \\
Ample:              & $\left|\frac{C_{n}^{\mathrm{af}}}{C_{n}^{\mathrm{af}}+C_{n}^{\mathrm{nf}}}-\frac{C_{n}^{\mathrm{as}}}{C_{n}^{\mathrm{as}}+C_{n}^{\mathrm{ns}}}\right|$                                                                                           & Jaccard:            & $\frac{C_{n}^{a f}}{C_{n}^{a f}+C_{n}^{n f}+C_{n}^{a s}}$                 \\ \\
Wong3:              & \multicolumn{3}{c}{$C_{n}^{\mathrm{af}}-h$ $\begin{array}{c}h = \left\{\begin{array}{ll}C_{n}^{a s}  &  \text { if } C_{n}^{a s} \leq 2 \\2+0.1\left(C_{n}^{a s}-2\right)  &  \text { if } 2<C_{n}^{a s} \leq 10 \\2.8+0.01\left(C_{n}^{a s}-10\right)  &  \text { if } C_{n}^{a s}>10\end{array}\right.\end{array}$}       \\
\bottomrule
\end{tabular}
}

\end{table*}

We then rank the quantitative metrics of these neurons from largest to smallest based on certain weights, with higher metrics indicating more suspicious neurons and the ones we needed to target for repair.

\subsection{Constraints-solving based Repairing}
\label{sec:repair_constraints}

After the neuron importance evaluation, for each layer, we obtain a vector of neuron importance. We rank this importance vector. The neuron with the highest importance is our target for repair as it could have the greatest impact on the corrected error outcome.

The optimization problem for a single neuron can be described as follows:
\begin{equation}
\label{eqn:LPProblem}
\begin{aligned}
& \textbf{Minimize:} \quad M \\
& \textbf{Subject to:} \\
& \quad M \geq 0 \\
& \quad \delta_i \in [-M, M] \quad \forall i \in \{1, 2, \ldots, n\} \\
& \\
& \textbf{If floating model gives the result 1 and quantized model gives 0:} \\
& \quad \forall x_i \text{ in TestSet } X: \sum_{i=1}^{m} w_i x_i < 0 \text{ and } \sum_{i=1}^{m} (w_i + \delta_i) x_i > 0 \\
& \\
& \textbf{If floating model gives the result 0 and quantized model gives 1:} \\
& \quad \forall x_i \text{ in TestSet } X: \sum_{i=1}^{m} w_i x_i > 0 \text{ and } \sum_{i=1}^{m} (w_i + \delta_i) x_i < 0 
\end{aligned}
\end{equation}

In the formula, $m$ represents the number of neurons connected to the previous layer of the selected neuron, and we number them from 1 to $m$. We add incremental $\delta$ to the weights to indicate the weights that need to be modified all the way to $m$. $M$ is used to make $\delta_{1} ... \delta_{i}$ are sufficiently small. The value $\delta_{1} ... \delta_{i}$ are encoded as the non-deterministic variables, and our task is to use Gurobi to solve these non-deterministic based on the given constraints.

We assume that in the full-precision neural network, this neuron's activation function gives the result $1$, and the quantized gives $0$. The corrected neuron in the quantized model result needs to be greater than 0 for the output of the activation function to be 1. If in the full-precision neural network, this neuron's activation function gives the result $0$, and the quantized gives $1$. The corrected neuron in the quantized model result needs to be smaller than 0 for the output of the activation function to be 0. In this case, we make the distance of the repaired quantized neural network as close as possible to that of the original quantized neural network.

The inputs to our algorithm are a quantized neural network $Q$ that needs to be repaired, a set of data sets $X$ for testing, and the full-precision neuron network model $F$ to be repaired.
We use Gurobi~\cite{optimization2012inc} as the constraint solver to solve the constraint and then replace the original weights with the result obtained as the new weights.

\subsection{\qnnrepair Algorithm}
\label{sec:algorithm}

Our repair method is formulated in Algorithm~\ref{alg:Repair}. The input to our algorithm is the full-precision model $F$, the quantized model $Q$. The repair set $X$, the validation set $V$, and the number of neurons that need to be repaired $N$ (Line 1). Firstly, we initialize arrays to store the activation states of the floating and quantized model, the values of the neuron importance, and four arrays $C^{a s}[]$, $C^{a f}[]$, $C^{n s}[]$ and $C^{n f}[]$ mentioned in Section~\ref{ranking_importance} (Line 1). For these six arrays, we set all elements to $0$.

Next, in lines 3-4, for each input in the test set $x \in \mathcal{X}_n$, we perform the inference process once obtain the neurons' activation states in the corresponding model layers and store them in the activation states of the floating and quantized model. In line 5, if $x[i]$ is a failing test, then we add the difference of activation status between the float model and quantized model to $C^{a s}[i]$, and vice versa. In line 11 and 12, we calculate $C^{n s}[]$ and $C^{n f}[]$ according to the definition in Section~\ref{ranking_importance}. We calculate the importance (here we use DStar as an example) for each neuron regarding seven importance metrics and sort them in descending order then store them in set $I_{n}[]$ in line 14. 

Then, we pick the neuron in $I_{n}[]$, according to the neuron's weights and the corresponding inputs from the previous layer, we create and solve the LP problem we discussed in Section \ref{sec:repair_constraints}, get the correction of each neuron, and update their weights. When it arrives at the maximum number of neurons to repair, the loop breaks and we have corrected all the neurons. These are implemented at lines 17-24 in Algorithm~\ref{alg:Repair}.  

\begin{algorithm}[H]
\resizebox{0.7\textwidth}{!}{%
\begin{minipage}{\linewidth}
\label{alg:Repair}
\LinesNumbered
\SetAlgoLined
\KwIn{Floating-point model $F$, Quantized model $Q$, Repair set $X$, Validation set $V$, Number of neurons to be repaired $N$}

\KwOut{Repaired model $Q'$, Repaired model's accuracy $Acc$}

Initialize $F_a[ ][ ], Q_a[ ][ ], I_n[ ]$, $C^{a s}[ ], C^{a f}[ ]$, $C^{n s}[ ], C^{n f}[ ]$ 

\ForEach{$X$}{
    $F_a[ ][ i] = \text{getActStatus}(F, x_i)$
    
    $Q_a[ ][ i] = \text{getActStatus}(Q, x_i)$
    
    \eIf{$x[i]$ is a failing test}{
        $C^{a f}[ i] = C^{a f}[ i] + |F_a[ ][ i] - Q_a[ ][ i]|$
    }{
        $C^{a s}[ i] = C^{a s}[ i] + |F_a[ ][ i] - Q_a[ ][ i]|$
    }
}

$C^{n f}[ ] = C^{n f}[ ] - C^{a f}[ ]$

$C^{n s}[ ] = C^{n s}[ ] - C^{a s}[ ]$

$I_n[ ] = \text{DStar}(C_n^{a s}[ ], C_n^{a s}[ ], C_n^{a s}[ ], C_n^{a s}[ ])$

$I_n[ ] = \text{sort}(I_n[ ])$ \tcp{In descending order}

Initialize weight of neurons $w[ ][ ]$ and the increment $\delta[ ][ ]$

\ForEach{$neuron[i] \in I_n[ ]$}{
    \ForEach{$edge[j][i] \in neuron[i]$}{
        $w[j][i] = \text{getWeight}(edge[j][i])$
    }
    $\delta[ ][ i] = \text{solve}(X, w[ ][ i])$ \tcp{Solve LP problem \ref{eqn:LPProblem}}
    \ForEach{$edge[j][i] \in neuron[i]$}{
        $edge[j][i] = \text{setWeight}(w[j][i] + \delta[j][i])$
        
        $Q' = \text{update}(Q, edge[j][i])$
    }
    \If{$i>= N$}{
        break
    }
    
}
$Acc = \text{calculateAcc}(Q', V)$

\Return{$Q'$}
\caption{Repair algorithm}
\end{minipage}%
}
\end{algorithm}

Finally, we evaluate the classification accuracy of the corrected quantized model. If it satisfies our requirements, then the model is repaired. Otherwise, try other combinations of parameters like important metrics or the maximum number of neurons needed to repair and repeat the LP solving and correction process. The output for this algorithm is the repaired model with updated weight.

\section{Experiment}
\label{sec:experiment}

\subsection{Experimental Setup}
\label{experiment_setup}

We conduct experiments on a machine with Ubuntu 18.04.6 LTS OS Intel(R) Xeon(R) Gold 5217 CPU @ 3.00GHz and two Nvidia Quadro RTX 6000 GPUs. The experiments are run with TensorFlow2 + nVidia CUDA platform. We use the Gurobi~\cite{optimization2012inc} as the linear program solver and enable multi-thread solving (up to 16 cores). We apply \qnnrepair to repair a benchmark of five quantized neural network models, including MobileNetV2~\cite{sandler2018mobilenetv2} on ImageNet datasets~\cite{deng2009imagenet}, and ResNet-18~\cite{he2016deep}, VGGNet~\cite{simonyan2014very} and two simple convolutional models trained on CIFAR-10 dataset~\cite{cifar10}. 
The details of these models are given in Table~\ref{table:baseline}.


\begin{table}[]
\centering
\caption{The baseline models. Parameters include the trainable and non-trainable parameters in the models; the unit is million (M). The two accuracy values are for the original floating point model and its quantized version, respectively.}
\label{table:baseline}

\begin{tabular}{llcccc}
\toprule
            &         &         &       &   \multicolumn{2}{c}{Accuracy}    \\ \cline{5-6} 
Model       & Dataset & \#Layers~~  & \#Params~~ & floating point~~ & quantized \\
\midrule
Conv3       & CIFAR-10 & 6       & 1.0M       & 66.48\%         & 66.20\%              \\
Conv5       & CIFAR-10 & 12      & 2.6M       & 72.90\%          & 72.64\%                    \\
VGGNet      & CIFAR-10 & 45      & 9.0M       & 78.67\%         & 78.57\%             \\
ResNet-18   & CIFAR-10 & 69      & 11.2M      & 79.32\%         & 79.16\%             \\
MobileNetV2~~ & ImageNet & 156     & 3.5M       & 71.80\%         & 65.86\% 
\\
\bottomrule
\end{tabular}
\end{table}

We obtained the full-precision MobileNetV2 directly from the Keras library, whereas we trained the VGGNet and ResNet-18 models on the CIFAR-10 dataset. We also defined and trained two smaller convolutional neural networks on CIFAR-10 for comparison: Conv3, which contains three convolutional layers, and Conv5, which contains five convolutional layers. Both models have two dense layers at the end. 
The quantized models are generated by using TensorFlow Lite (TFlite)~\cite{tflite} from the floating point models. In  TFLite, we chose dynamic range quantization, and the weights are quantized as 8-bit integers. The quantized convolution operation is optimized for performance, and the calculations are done in the fixed-point arithmetic domain to avoid the overhead of de-quantizing and re-quantizing tensors. 

For repairs of the quantized model's performance, we use a subset of ImageNet called ImageNet-mini~\cite{ifigotin}, which contains 38,668 images in 1,000 classes. The dataset is divided into the repair set and the validation set. The repair set contains 34,745 images, and the validation set contains 3,923 images. The CIFAR-10 dataset contains 60,000 images in 10 classes in total. 50,000 of them are training image, and 10,000 of them are test set. We use 1,000 images as the repair set.
We use the repair set to identify suspicious neurons, generate LP constraints, apply corrections to the identified neurons, and use the validation set to evaluate the accuracy of the models. We repeat the same experiment ten times for random neuron selection and get the average to eliminate randomness in repair methods. 

\subsection{Repair Results on Baselines}
\label{sec:experiment_results}


In this part, we apply \qnnrepair to these baseline quantized models, except for MobilenetV2, in Table \ref{table:baseline}. In our experiments, MobileNetV2 is trained on ImageNet while other models are trained on CIFAR-10, and it contains more layers. The results for MobileNetV2 are reported in Section \ref{sec:neuron_selection}.
For each model, we perform a layer-by-layer repair of its last dense layers. 
We name these dense layers dense-3 (the third last layer), dense-2 (the second last layer), and dense-1 (the output layer).

\begin{table*}
\centering
\caption{\qnnrepair results  on CIFAR-10 models. The best repair outcome for each model, w.r.t. the dense layer in that row, is in \textbf{bold}. We further highlight the best result in \boxed{\color{blue} blue} if the repair result is even better than the floating point model and in {\color{red} red} if the repair result is worse than the original quantized model. 
Random means that we randomly select neurons at the corresponding dense layer for the repair, whereas Fault Localization refers to the selection of neurons based on important metrics in \qnnrepair. In All cases, all neurons in that layer are used for repair. \lab{n/a} happens when the number of neurons in the repair is less than 100, and \lab{-}  is for repairing the last dense layer of 10 neurons, and the result is the same as the All case. }
\label{table:importance_layers}
\resizebox{\linewidth}{!}{
\begin{tabular}{@{\extracolsep{6pt}}cccccccccc@{}}
\hline
             & \multicolumn{4}{c}{Random}                                                                              & \multicolumn{4}{c}{Fault Localization}                                                                            & -                  \\ \cline{2-5} \cline{6-9}
\#Neurons repaired             & 1                           & 5                           & 10                   & 100              & 1                           & 5                           & 10                          & 100                     & All                    \\ \hline
Conv3\_dense-2                 & 63.43\%                     & 64.74\%                     & 38.90\%               & n/a              & 66.26\%                     & {\textbf{66.36}}\%                     & 62.35\%                     & n/a                      & 57.00\%                 \\
Conv3\_dense-1 & 65.23\%                     & 66.31\%                     & -                    & n/a                    & 66.10\%                      & 66.39\%                     & -                           & n/a                           & {\textbf{66.46}}\%              \\
Conv5\_dense-2                 & \multicolumn{1}{l}{72.49\%} & \multicolumn{1}{l}{72.55\%} & \multicolumn{1}{l}{72.52\%} & \multicolumn{1}{l}{72.52\%} & \multicolumn{1}{l}{{\color{red}\textbf{72.56}}\%} & \multicolumn{1}{l}{72.56\%} & \multicolumn{1}{l}{72.56\%} & \multicolumn{1}{l}{72.56\%} & \multicolumn{1}{l}{72.54\%} \\
Conv5\_dense-1                 & 72.51\%                     &  72.52\%                           & -                    & n/a                    & {\color{red}\textbf{72.58}}\%                     &       72.56\%                      & -                           & n/a                           &      72.56\%               \\
VGGNet\_dense-3                & 78.13\%                     & 78.44\%                     & 78.20\%               & 78.38\%              & \boxed{{\color{blue}\textbf{78.83}}\%}                     & 78.82\%                     & 78.78\%                     & 78.66\%                     & 78.60\%               \\
VGGNet\_dense-2                & 78.36\%                     & 78.59\%                     & 78.44\%              & 78.22\%              & 78.55\%                     & \boxed{{\color{blue}\textbf{78.83}}\%}                     & 78.83\%                     & 78.83\%                     & 78.83\%              \\
VGGNet\_dense-1                & 78.94\%                     & 67.75\%                     & -                    & n/a                    & \boxed{{\color{blue}\textbf{79.29}}\%}                     & 69.04\%                     & -                           & n/a                           & 74.49\%              \\
ResNet\_dense\_1               & 78.90\%                            &   78.92\%                         &   -                   & n/a                     & 79.08\%                           & {\textbf{79.20}}\%                          &  -                          & n/a                           &  78.17\%                    \\ \hline
\end{tabular}
}

\end{table*}

The \qnnrepair results are reported in Table \ref{table:importance_layers}. We ranked the neurons using important metrics and chose the best results among the seven metrics. We also run randomly picked repairing as a comparison. We have chosen Top-1, Top-5, Top-10, Top-100, and all neurons as the repairing targets. For most models, the repair improves the accuracy  of the quantized network, and in some cases, even higher than the accuracy of the floating-point model. 

The dense-2 layer only contains 64 neurons in the Conv3 model. Hence we selected 30 neurons as the repair targets. In the dense-1 layer of Conv3, the effect of repairing individual neurons is not ideal, but as the number of repaired neurons gradually increases, the more correct information the Conv3 quantization model obtains from the floating-point model, so the accuracy gradually improves until it reaches 66.46\% (which does not exceed the accuracy of the floating-point Conv3 neural network, but it gets very close to it: 66.48\%, see Table~\ref{table:baseline} ). This is because all the repair information in the last layer comes from the original floating-point neural network. Note that because of the simple structure of the Conv3 neural network, the floating-point version of Conv3 itself is inaccurate, and the quantized and repaired neural network does not exceed the accuracy. In the dense-2 layer of Conv5, applying importance metrics to repair this layer is slightly better than random selection, only 0.01\% regarding randomly selecting 5 neurons compared with using fault localization to select Top-5 neurons. Compared to the quantized model before repair, whose accuracy is 72.64\%, the repairing only gets an accuracy of 72.56\%, which does not improve the model's accuracy. In the dense-1 layer of Conv5, the best result is using fault localization to pick the Top-1 neuron and repair at 72.58\% accuracy, and this is not better than the quantized model before repair.

For VGGNet and ResNet-18 neural networks, the dense-1 layer is a good comparison. Both VGGNet and ResNet-18 have relatively complex network structures, and the accuracy of the original floating-point model is close to 80\%. In the dense-1 layer of ResNet-18, only some of the neurons were repaired with accuracy close to their original quantized version, but all of them did not exceed the exact value of the floating-point neural network after the repair. However, unlike ResNet-18, correcting a single neuron randomly in the dense-1 layer of VGGNet make it more accurate than the quantized version of VGGNet. Using the importance metric and correcting a single neuron make the accuracy even higher than the floating-point version of VGGNet. However, repairing dense-1 of VGGNet was unsatisfactory, especially when 5 neurons are selected for repair; it suffered a significant loss of accuracy, even below 70\%, which was regained if all ten neurons in the last layer were repaired. In the dense-2 layer of VGGNet, the overall accuracy is higher than 78\%. When the importance metric is applied, the accuracy reaches 78.83\%, noting that this accuracy is also achieved if all neurons in this layer are repaired. For the dense-3 layer of VGGNet, repairing 5 or 10 neurons using importance metrics will achieve the highest accuracy at 78.83\%, the same as repairing the dense-2 layer. 



\begin{table*}[]
\centering
\caption{\qnnrepair results on ImageNet model. }
\resizebox{0.7\linewidth}{!}{
\begin{tabular}{@{\extracolsep{6pt}}cccccc@{}}
\toprule
   & \multicolumn{2}{c}{Random} & \multicolumn{2}{c}{Fault Localization} & -     \\
\cmidrule(lr){2-3}   
\cmidrule(lr){4-5}  
\#Neurons repaired   & 10           & 100         & 10                   & 100                 & All       \\ \hline
MobileNetV2\_dense-1 & 70.75\%      & 70.46\%     & {\textbf{70.77}}\%             & 70.00\%             & 68.98\%\\
\bottomrule
\end{tabular}
}
\end{table*}

\paragraph{ImageNet} We also conducted repair on the last layer for MobileNetV2 trained on the ImageNet dataset of high-resolution images. Using Euclid as the importance metric and picking 10 neurons as the correct targets achieve the best results, at 70.77\%, improving the accuracy the quantized model. 

\subsection{Comparison with Data-free Quantization}

We tested SQuant~\cite{guo2022squant}, a fast and accurate data-free quantization framework for convolutional neural networks, employing the constrained absolute sum of error (CASE) of weights as the rounding metric. We tested SQuant two quantized models, the same as our approach: MobileNetV2 trained on ImageNet and ResNet-18 on CIFAR-10. We made some modifications to the original code to support MobileNetV2, which is not reported in their experiments.

\begin{table*}[]
\centering
\caption{\qnnrepair vs SQuant}
\label{table:compare_sota}
\resizebox{0.6\textwidth}{!}{%
\begin{tabular}{lcccc}
\toprule
                    & \multicolumn{2}{c}{MobileNetV2}          & \multicolumn{2}{c}{ResNet-18}           \\
\cmidrule(lr){2-3}   
\cmidrule(lr){4-5} 
\multicolumn{1}{l}{}      & \multicolumn{1}{l}{Accuracy} & Time      & \multicolumn{1}{l}{Accuracy} & Time     \\
\midrule
SQuant~\cite{guo2022squant}                    & 46.09\%                     & 1635.37ms & 70.70\%                     & 708.16ms \\
\qnnrepair & \textbf{70.77\%}                      & $\sim$15h & \textbf{79.20\%}                       & $\sim$9h  \\
\bottomrule
\end{tabular}%
}
\end{table*}

In contrast, to complete data-free quantization, our constraint solver-based quantization does not require a complete dataset but only some input images for repair. Despite taking much more time than SQuant because it uses Gurobi and a constrained solution approach, MobileNetV2 -- a complex model trained on ImageNet -- \qnnrepair achieves much higher accuracy.

\subsection{Repair Efficiency}

The constraints-solving part contributes to the major computation cost in \qnnrepair.
For other operations, such as importance evaluation, modification of weights, model formatting, etc., it takes only a few minutes to complete.
Thereby, Table~\ref{table:running_time} measures the runtime cost when using the Gurobi to solve the values of the new weights for a neuron for our experiments on the VGGNet model. 
It is shown in Table~\ref{table:running_time} that 75\% of the solutions were completed within 5 minutes, and less than 9\% of the neurons could not be solved, resulting in a total solution time of 9 hours for a layer of 512 neurons. 

\begin{table*}[]
\centering
\caption{The Gurobi solving time for constraints of each neuron in the dense-2 layer of the VGGNet model. There are 512 neurons in total.}
\label{table:running_time}
\begin{tabular}{@{\extracolsep{5pt}}cccccc@{}}
\toprule
Duration   & \textless{}=5mins & 5-10mins & 10-30mins & 30mins-1h & No solution \\
\midrule
Percentage & 75\%              & 8.98\%   & 5.27\%    & 1.76\%    & 8.98\%          \\
\bottomrule
\end{tabular}
\end{table*}

\subsection{Comparison Between Fault Localization Metrics in \qnnrepair}
\label{sec:neuron_selection}

We let the model and the layer stay the same. We use MobileNetV2 and the last layer as our target. We compare seven representative important metrics mentioned in Section~\ref{sec:neuron_selection}. In these experiments, we used 1,000, 500, 100, and 10 jpeg images as the repair sets to assess the performance of different importance assessment methods.

\begin{figure*}[htbp]
\centering
\begin{minipage}{\textwidth}
\begin{subfigure}[b]{0.5\textwidth}
\includegraphics[width=\textwidth, clip, trim=8mm 8mm 8mm 8mm]{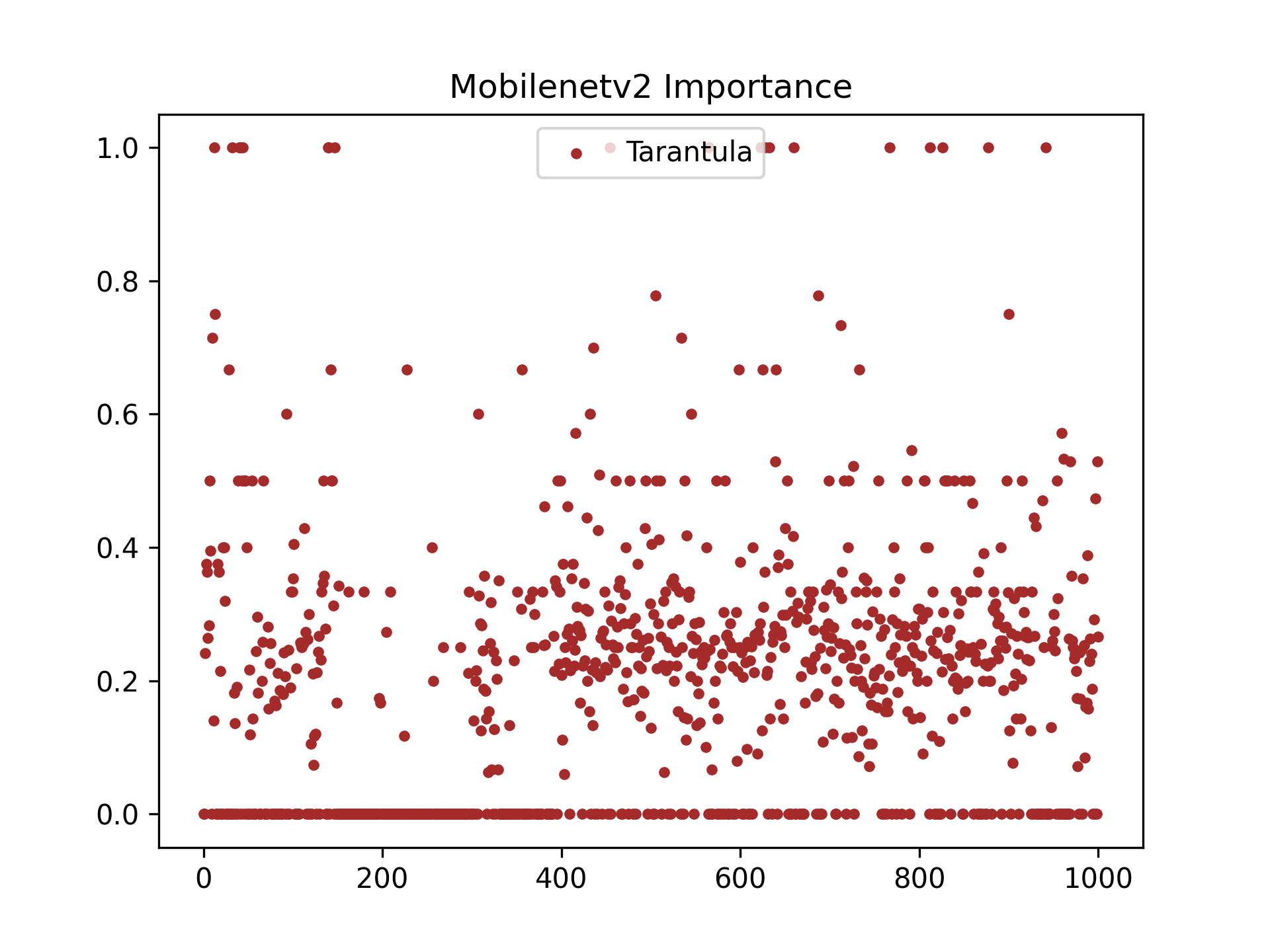}
\end{subfigure}
\hfill
\begin{subfigure}[b]{0.5\textwidth}
\includegraphics[width=\textwidth, clip, trim=8mm 8mm 8mm 8mm]{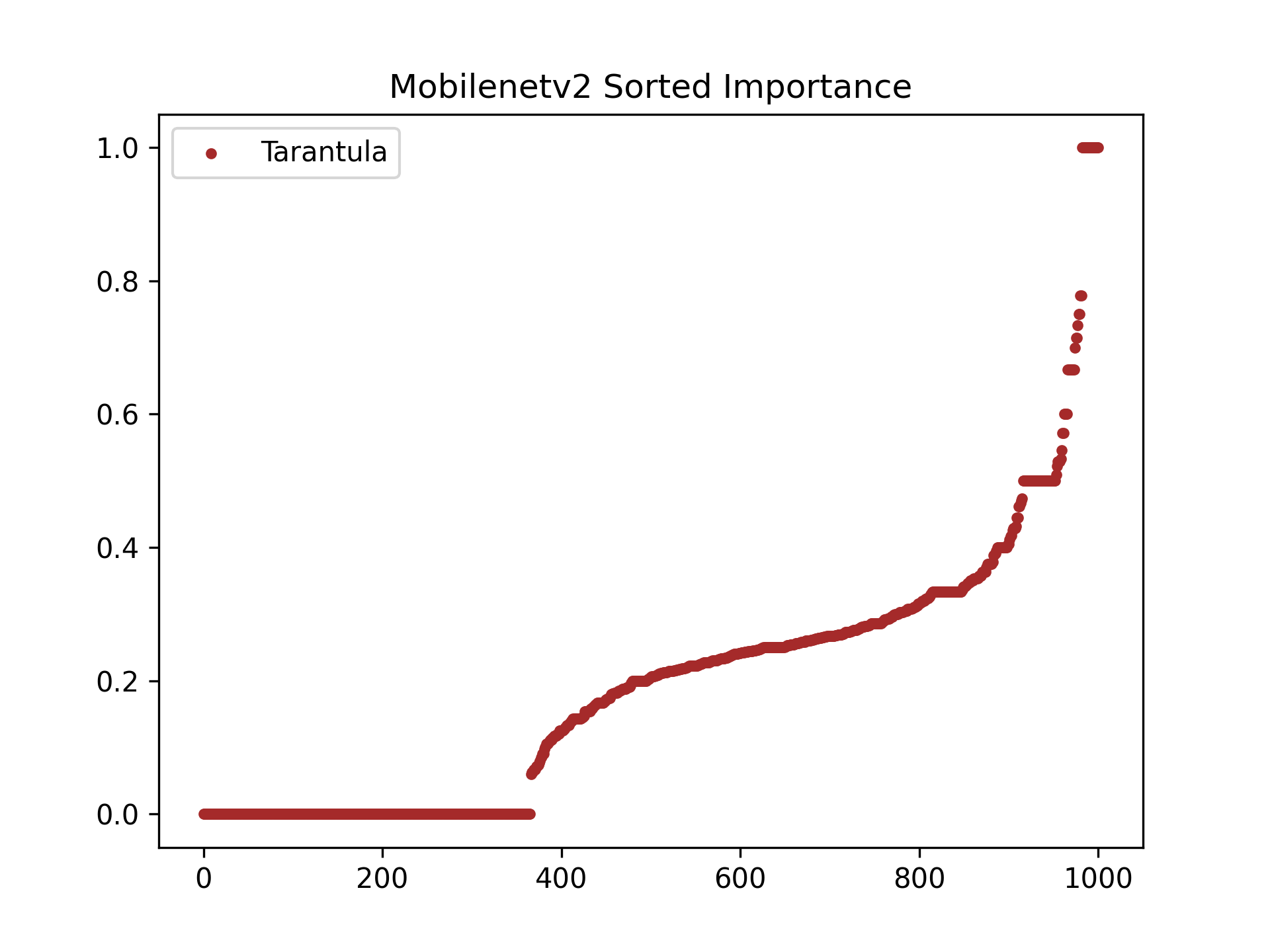}
\end{subfigure}
\hfill
\end{minipage}
\caption{Importance distribution regarding certain importance metrics on MobileNetV2.}
\label{fig:combined}
\end{figure*}

Firstly, we rank the neurons in the last layer using seven different representative important metrics, which are Tarantula~\cite{jones2005empirical}, Ochiai~\cite{abreu2007accuracy}, DStar~\cite{wong2013dstar}, Jaccard~\cite{agarwal2014fault}, Ample~\cite{dallmeier2005lightweight}, Euclid~\cite{galijasevic2002fault} and Wong3~\cite{wong2007effective}. As shown in Figure~\ref{fig:combined}, for the last fully connected layer of MobileNetV2, the important neurons are mainly concentrated at the two ends, those neurons with the first and last numbers. The evaluation metrics results are relatively similar for different neurons.

\begin{table*}
\centering
\caption{The results regarding importance metrics, including 7 fault localization metrics and 1 random baseline. The number of images indicates how many inputs are in the repair set. 
}
\label{table:importance_metrics}
\resizebox{\linewidth}{!}{
\begin{tabular}{@{\extracolsep{6pt}}llllllllll@{}}
\toprule Model+Repair Layer & \#Images & Tarantula & Ochiai   & DStar    & Jaccard  & Ample    & Euclid   & Wong3    & Random    \\
\midrule
MobileNetV2\_dense-1 & 1000      & 70.61\%                             & 69.76\%                                                                                                                                      & 69.73\%                                                                  & 69.73\%                           & 69.72\%                         & {\textbf{70.70}}\%                                         & 69.73\%                & 69.56\% \\
  & 500       & 68.99\%                             & 69.01\%                                                                                                                                      & 69.05\%                                                                   & 69.05\%                            & 68.99\%                         & {\textbf{69.46}}\%                                         & 69.06\%                & 69.00\% \\
 & 100       & 69.50\%                             & 69.42\%                                                                                                                                      & 69.46\%                                                                  & 69.46\%                           & 69.53\%                         & 69.98\%                                         & 69.46\%                & {\textbf{70.12}}\% \\
  & 10        & 70.62\%                              & 70.15\%                                                                                                                                      & 70.12\%                                                                   & 70.12\%                            & 70.17\%                         & {\textbf{70.73}}\%                                          & 70.12\%                 & 70.18\%  \\ \hline
VGGNet\_dense-3 & 1000        & 78.64\%                              & 78.64\%                                                                                                                                       & 78.64\%                                                                   & 78.64\%                            & 78.65\%                          & {\textbf{78.66}}\%                                          & {\textbf{78.66}}\%                 & 78.22\%  \\
VGGNet\_dense-2 & 1000          & {\textbf{78.83}}\%                              & {\textbf{78.83}}\%                                                                                                                                       & {\textbf{78.83}}\%                                                                   & {\textbf{78.83}}\%                            & {\textbf{78.83}}\%                         & {\textbf{78.83}}\%                                          & {\textbf{78.83}}\%                 & 78.38\%  \\
Conv3\_dense & 1000           & {\textbf{59.50}}\%                              & {\textbf{59.50}}\%                                                                                                                                       & {\textbf{59.50}}\%                                                                    & {\textbf{59.50}}\%                             & 59.27\%                          & 59.27\%                                          & 59.27\%                 & 32.42\% 
\\
\bottomrule
\end{tabular}
}
\end{table*}


We selected the 100 neurons (for Conv3, it is 30 neurons) with the highest importance and could be solved by MILP solvers according to different importance measures. The deltas are obtained according to Equation~\ref{eqn:LPProblem}, and we apply the deltas to the quantized model. After that, we use the validation sets from ImageNet, which contains 50,000 jpeg images, to test the MobileNetV2 model. We also use the validation sets from CIFAR-10, which contains 10,000 png image files, to test VGGNet and Conv5 after the repair. As a comparison, we also randomly picked 100 neurons to apply to repair and tested their accuracy. We give the results of the top 100 important neurons after selection and repair, as shown in Table~\ref{table:importance_metrics}.

We pick Tarantula and plot the scatter plots based on the importance distribution of the different neurons. We rank the importance of those neurons and draw line plots as illustrated in Figure~\ref{fig:combined}.

The figures give scatter plots of neuron importance and ranked line plots for the last dense layer of MobileNetV2. The horizontal coordinates of these plots are the serial numbers of the neurons. For the last layer in the MobileNetV2 model, few neurons have the highest importance. More than 300 neurons had an importance measurement of 0, and another large proportion had an importance of 0.5 or less. Based on the ranking of the importance of neurons, all the evaluation metrics except Tarantula and Euclid considered 108, 984, 612, 972 to be the four most important neurons in this layer, and among the 5th-10th most important neurons, 550, 974, 816, 795 and 702, just in a different order. This is reflected in the importance distribution graphs as spikes at the ends and as spikes at the ends of the graphs. Hence Ochiai, Dstar, Jaccard, Ample, and Wong3 have similar performance regarding the accuracy evaluation, and Euclid and Tarantula achieve better accuracy on ImageNet validation sets. 

Table~\ref{table:importance_metrics} shows that the Euclid importance assessment method is highly effective, achieving relatively good results from restoration with 500 images to restoration with ten images and achieving only weaker accuracy than the Tarantula method in a restoration scenario with 1,000 images. A random selection of neurons can achieve good restoration results, especially when we select 100 images as restoration images,  it 
has a validation accuracy higher than 70\%. Also, in Table~\ref{table:importance_metrics}, our methods work well with the models containing fewer neurons. In experiments with Conv3\_dense, our approach achieves more than 20\% higher accuracy than random selection. When it comes to large models, although it is not as obvious as smaller models, but still has higher accuracy than random selection in most cases, even if random selection is better than importance ranking, which is only a little bit better (0.14\%). Considering the successful and failing tests used for repair, i.e., the repair images, in our experiments, the repair results of using 10 repair images were slightly better than using 1,000 repair images. For the Euclid method that produces the best repair results, the accuracy of using 10 images is 0.03\% higher than using 1,000 repairing images.

For the VGGNet model, for the same reason as MobileNetV2 regarding the neuron importance ranking, the Tarantula, Ochiai, DStar, Jaccard, Euclid, and Wong3 give the same results when selecting 100 top important neurons to repair. As a comparison, the accuracy of random selection in dense-2 layer and dense-3 has a slight drop, at 78.38\% and 78.22\%. For Conv3 model, the seven importance metrics give the same results, and randomly selected 30 neurons suffered a great accuracy loss, at 32.42\%. But compared to the results in Table~\ref{table:importance_layers}, repairing 30 top neurons also suffered accuracy drops. For the dense layer of conv3, the best repair is still to select one neuron for repair based on Tarantula sorting at 66.10\%, and if random selection is taken into account, then selecting five neurons for repair would give the best result at 64.74\%. 

We also conducted a side-by-side comparison of the number of images required for the repair on MobileNetV2. It shows that the best results are obtained using 1,000 images for repair and 10 images for the repair, but given the amount of time required to generate constraints for the repair using 1,000 images and to solve the constraints using Gurobi, we recommend using a smaller set of repair images for the model.

Euclid demonstrates that it has the highest accuracy most of the time, and repairing with importance evaluation is more accurate than repairing randomly selected neurons. 


\subsection{Limitations}

According to Nemhauser and Wolsey~\cite{nemhauser1988integer}, the MILP problem is NP-Hard. There is no known polynomial time algorithm that can solve all MILP instances. Therefore, for very large or structurally complex problems, the solver may take a very long time to find the optimal solution or an acceptable approximate solution. Hence, selecting more repairing images for correction will have a greater likelihood of Gurobi being unable to solve the MILP problem, reflected in the limitation of improving accuracy.

\section{Conclusion}
\label{Conclusion}

In this paper, we presented \qnnrepair, a novel method for repairing quantized neural networks. Our method is inspired by traditional software statistical fault localization. We evaluated the importance of the neural network models and used Gurobi to get the correction for these neurons. According to the experiment results, after correcting the model, accuracy increased compared with the quantized model. We also compared our method with state-of-the-art techniques; the experiment results show that our method can achieve much higher accuracy when repair models are trained on large datasets.


As the future works, we will move forward to larger datasets; currently, we support MobileNetV2 trained on ImageNet. In the future, we will test our tool and make it scalable for larger models and not limited to classification tasks like GPT and stable diffusion. For these large networks, due to the complexity of the model itself, repairing them will require a lot of computational resources, and we will find a balance between improving accuracy and computing time.

For some of the repairing problems, Gurobi was not able to solve them in the given time limit, so in the future, we intend to optimize the encoding of the neural network repair problem to increase the speed of the repair solution and to solve some of the repair problems that were not previously solved. We will also try more problem solvers in the future, such as SMT solvers, to solve these problems that Gurobi cannot solve.

\section*{Acknowledgements}

This work is funded by the EPSRC grants EP/T026995/1, EP/V000497/1, EU H2020 ELEGANT 957286, Soteria project awarded by the UK Research and Innovation for the Digital Security by Design (DSbD) Programme, and Cal-Comp Electronic by the R\&D project of the Cal-Comp Institute of Technology and Innovation.
\newpage
\balance
\bibliographystyle{splncs04}
\bibliography{references}

\end{document}